\def\year{2021}\relax
\documentclass[letterpaper]{article} 
\usepackage{aaai21}  
\usepackage{times}  
\usepackage{helvet} 
\usepackage{courier}  
\usepackage[hyphens]{url}  
\usepackage{graphicx} 
\usepackage{natbib}  
\usepackage{caption} 
\usepackage{xspace}
\usepackage{subfigure}
\usepackage{amssymb}
\usepackage{amsmath}
\usepackage{todo}
\usepackage{comment}
\usepackage{bm}
\usepackage{color}
\usepackage{multirow}
\usepackage{booktabs}
\usepackage{graphicx}

\usepackage{xcolor}
\usepackage[normalem]{ulem} 
\newcommand\hl{\bgroup\markoverwith
  {\textcolor{yellow}{\rule[-.5ex]{2pt}{2.5ex}}}\ULon}
\newcommand\tl{\bgroup\markoverwith
  {\textcolor{red}{\rule[-.5ex]{2pt}{2.5ex}}}\ULon}
\newcommand\rl{\bgroup\markoverwith
  {\textcolor{green}{\rule[-.5ex]{2pt}{2.5ex}}}\ULon}

\newcommand\shorteq{\vspace{-2pt}}

\newcommand\ReorderNAT{ReorderNAT\xspace}
\newcommand\shrink{\vspace{-1em}}

\title{Guiding Non-Autoregressive Neural Machine Translation Decoding \\with Reordering Information}

\author{Qiu Ran\thanks{indicates equal contribution}, Yankai Lin\footnotemark[1], Peng Li\footnotemark[1], Jie Zhou\\}

\affiliations{Pattern Recognition Center, WeChat AI, Tencent Inc., China\\
\{soulcaptran,yankailin,patrickpli,withtomzhou\}@tencent.com}

\date{}

\begin{document}
\maketitle
\begin{abstract}
Non-autoregressive neural machine translation (NAT) generates each target word in parallel and has achieved promising inference acceleration. However, existing NAT models still have a big gap in translation quality compared to autoregressive neural machine translation models due to the multimodality problem: the target words may come from multiple feasible translations. To address this problem, we propose a novel NAT framework \ReorderNAT which explicitly models the reordering information to guide the decoding of NAT. Specially, \ReorderNAT utilizes deterministic and non-deterministic decoding strategies that leverage reordering information as a proxy for the final translation to encourage the decoder to choose words belonging to the same translation. Experimental results on various widely-used datasets show that our proposed model achieves better performance compared to most existing NAT models, and even achieves comparable translation quality as autoregressive translation models with a significant speedup.
\end{abstract}

\section{Introduction}
\label{sec:introduction}


Neural machine translation (NMT) models with encoder-decoder framework~\citep{sutskever2014sequence,bahdanau2014neural} significantly outperform conventional statistical machine translation models~\citep{koehn2003statistical,koehn2007moses}.
Despite their success, the state-of-the-art NMT models usually suffer from the slow inference speed, which has become a bottleneck to apply NMT in real-world translation systems. The slow inference speed of NMT models is due to their autoregressive property, i.e., decoding the target sentence word-by-word according to the translation history.

Recently, \citet{gu2018non} introduced non-autoregressive NMT (NAT) which can simultaneously decode all target words to break the bottleneck of the autoregressive NMT (AT) models. To this end, NAT models~\citep{gu2018non,wei2019imitation,wang2019non,guo2019non} usually directly copy the source word representations to the input of the decoder, instead of using previous predicted target word representations. Hence, the inference of different target words are independent, which enables parallel computation of the decoder in NAT models. NAT models could achieve $10$-$15$ times speedup compared to AT models while maintaining considerable translation quality.

However, NAT models still suffer from the multimodality problem~\cite{gu2018non}: it discards the dependencies among the target words, and therefore the target words may be chosen from multiple feasible  translations, resulting in duplicate, missing or even wrong words. 
For example, the German phrase ``{\it Vielen Dank}'' can be translated as both ``\hl{{\it thank you}}'' and ``\rl{{\it many thanks}}''. Unfortunately, as each target word is generated independently, ``{\it \hl{thank} \rl{thanks}}'' and ``{\it \rl{many} \hl{you}}'' may also be assigned high probabilities, resulting in inferior translation quality. 
In this work, we argue reordering information is essential for NAT models and helpful for alleviating the multimodality problem.

\begin{figure*}
    \centering
    \includegraphics[width=0.9\textwidth]{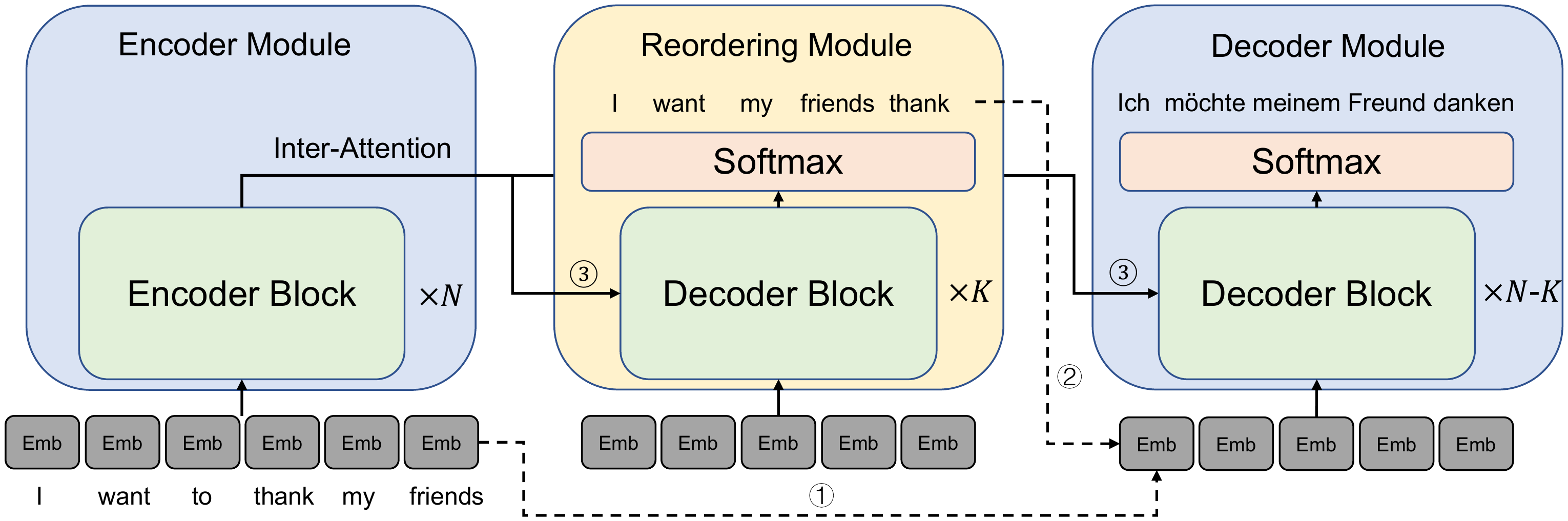}
    \caption{The architecture of our \ReorderNAT model. Different from original NAT models, our model adds a reordering module between the encoder and decoder modules to explicitly model the reordering information. For original NAT models, the decoder inputs are the copied embeddings of source sentence (No.1 dashed arrow), and for our \ReorderNAT model, the decoder inputs are the embeddings of pseudo-translation generated by reordering module (No. 2 dashed arrow).
    The encoder and decoder blocks are the same as existing NMT models (e.g., Transformer block).}
    \label{fig:model} 
    \shrink
 \end{figure*}

To this end, we propose a novel NAT framework named \ReorderNAT in this work, which explicitly models the reordering information to guide the decoding of NAT. To be specific, as shown in Figure~\ref{fig:model}, \ReorderNAT first reorders the source sentence into a pseudo-translation formed by source words but in the target language word order, and then translates the source sentence conditioned on it. 
We further introduce two guiding decoding strategies which utilize the reordering information (i.e. pseudo-translation) to guide the word selection in decoding. The first one is {\em deterministic guiding decoding} which first generates a most likely pseudo-translation and then generates the target sentence based on it. The second one is {\em non-deterministic guiding decoding} which utilizes the conditional distribution of the pseudo-translation as a latent variable to guide the decoding of target sentences.

Ideally, the pseudo-translation can be viewed as a final translation written in source language. Guiding decoding with it could help to model the conditional dependencies of the target words and encourage the decoder to choose words belonging to the same translation, which naturally reduces the multimodality problem.
Moreover, the decoding space of generating pseudo-translation is limited to the permutation of words in the source sentence, which can be well modeled by a small model. Therefore, \ReorderNAT could effectively alleviate the multimodality problem by introducing the reordering information in NAT.



Experimental results on several widely-used benchmarks show that our proposed \ReorderNAT model achieves significant and consistent improvements compared to existing NAT models by explicitly modeling the reordering information to guide the decoding. Moreover, by introducing a simple but effective AT module to model reordering information, our \ReorderNAT immensely narrows the translation quality gap between AT and NAT models, while maintaining considerable speedup (nearly six times faster). The source codes are available at \url{https://github.com/ranqiu92/ReorderNAT}.

\section{Background}
\label{sec:background}

Non-autoregressive neural machine translation (NAT) is first proposed by~\citet{gu2018non} to alleviate the slow decoding issue of autoregressive neural machine translation (AT) models, which could simultaneously generate target words by removing their dependencies. Formally, given a source sentence $\mathbf{X}=\{x_1, \cdots, x_n\}$ and a target sentence $\mathbf{Y} = \{y_1, \cdots, y_m\}$, NAT models the translation probability from $\mathbf{X}$ to $\mathbf{Y}$ as a product of conditionally independent target word probability:
\shorteq
\begin{equation}
\label{eq:nat}
    \vspace{-0.1em}
    P(\mathbf{Y}|\mathbf{X}) = \prod_{i=1}^m P(y_i|\mathbf{X}).
    \vspace{-0.9em}
\end{equation}
\shorteq

Instead of utilizing the translation history, NAT models usually copy source word representations as the input of the decoder. Hence, when translating a sentence, NAT models could predict all target words with their maximum likelihood individually by breaking the dependency among them, and therefore the decoding procedure of NAT models is in parallel and has very low translation latency.

However, since NAT models discard the sequential dependencies among words in the target sentence, they suffer from the potential performance degradation due to the multimodality problem. To be specific, a source sentence may have multiple translations. During decoding, NAT models may choose the target words from different translations, resulting in duplicate, missing or even wrong words. Consequently, NAT models cannot effectively learn the intricate translation patterns from source sentences to target sentences, leading to inferior translation quality.

\section{Methodology}
\label{sec:methodology}

In this section, we introduce a novel NAT model named \ReorderNAT, which aims to alleviate the multimodality problem in NAT models.

\subsection{\ReorderNAT}
As shown in Figure~\ref{fig:model}, \ReorderNAT employs a reordering module to explicitly model the reordering information in the decoding\footnote{We do not employ positional attention~\citep{gu2018non} as the mechanism may be misguided by target supervision due to the indirect optimization and lead to inferior translation.}. Formally, \ReorderNAT first employs the reordering module to translate the source sentence $\mathbf{X}$ into a pseudo-translation $\mathbf{Z} = \{z_1, \cdots, z_m\}$ which reorganizes source sentence structure into the target language, and then uses the decoder module to generate target translation $\mathbf{Y}$ based on the pseudo-translation. \ReorderNAT models the overall translation probability as:
\begin{eqnarray}
    P(\mathbf{Y}|\mathbf{X}) &=& \sum_{\mathbf{Z}}P(\mathbf{Y}|\mathbf{Z},\mathbf{X})P(\mathbf{Z}|\mathbf{X}),
    \shorteq
\end{eqnarray}
where $P(\mathbf{Z}|\mathbf{X})$ is modeled by the reordering module and $P(\mathbf{Y}|\mathbf{Z},\mathbf{X})$ is modeled by the decoder module. Next, we will introduce the reordering and decoder modules in detail\footnote{The encoder module of \ReorderNAT is a multi-layer Transformer~\cite{vaswani2017attention}, which is the same as original NAT models.}.

\subsubsection{Reordering Module}

The reordering module determines the source-side information of each target word by learning to translate the source sentence into a pseudo-translation. We propose two feasible implementations of the reordering module:

 (1) \textbf{NAT Reordering Module}: Intuitively, the pseudo-translation probability can also be modeled as NAT:
\shorteq
\begin{equation}
\label{eq:word-reorder}
    P(\mathbf{Z}|\mathbf{X}) = \prod_{i=1}^m P(z_i|\mathbf{X}),
\shorteq
\end{equation}
where $P(z_i|\mathbf{X})$ is calculated by a single-layer Transformer. During inference, the NAT reordering module needs to determine the length of the pseudo-translation. To this end, we use a length predictor and copy the embeddings of the source sentence as the input of the reordering module similar to existing NAT models. 

(2) \textbf{AT Reordering Module}: We find that AT models are more suitable for modeling the reordering information compared to NAT models, and even a light AT model with similar decoding speed to a large NAT model could achieve better performance in modeling reordering information. Hence, we also introduce a light AT model to model the pseudo-translation probability as: 
\shorteq
\begin{equation}
\label{eq:word-reorder1}
    P(\mathbf{Z}|\mathbf{X}) = \prod_{i=1}^m P(z_i|\mathbf{z}_{<i},\mathbf{X}),
    \shorteq
\end{equation}
where $\mathbf{z}_{<i} = \{z_1, \cdots, z_{i-1}\}$ indicates the pseudo-translation history, and $P(z_i|\mathbf{z}_{<i},\mathbf{X})$ is calculated by a single-layer recurrent neural network.


\subsubsection{Decoder Module}

The decoder module translates the source sentence into the target translation with the guiding of  pseudo-translation, which regards the translation of each word as NAT:
\shorteq
\begin{equation}
\label{eq:word-translation}
    P(\mathbf{Y}|\mathbf{Z},\mathbf{X}) = \prod_{i=1}^m P(y_i|\mathbf{Z},\mathbf{X}).
    \shorteq
\end{equation}

As shown in Figure~\ref{fig:model}, the encoder module and  the decoder module can be viewed as a seq-to-seq model which translates the source sentence to target sentence. Different from original NAT, the input of our decoder module is the embeddings of pseudo-translation instead of copied embeddings of source sentence, which is used to guide the word selection. Note that the encoder outputs are also fed into the decoder attention module, which can help alleviate the reordering errors.

To make the model parameter number comparable with the baseline model, 
we use $K$ and $N-K$ decoder blocks for the reordering and decoder modules respectively
\footnote{We set $K$ to $1$ for an AT module while $N-1$ for an NAT module as it is more difficult for an NAT module to model the reordering information (see Experiments).}.


\subsection{Guiding Decoding Strategy}

\ReorderNAT explicitly models reordering information of NAT and aims to utilize it to alleviate the multimodality problem. Now the remaining problem is how to perform decoding with the guide of reordering information. We propose to utilize the pseudo-translation as a bridge to guide the decoding of the target sentence, which can be formulated as:
\shorteq
\begin{eqnarray}
\label{eq:split}
    \mathbf{Y}^*&=& \mathop{\arg\max}_{\mathbf{Y}}P(\mathbf{Y}|\mathbf{X}) \nonumber\\
    &=& \mathop{\arg\max}_{\mathbf{Y}}\sum_{\mathbf{Z}} P(\mathbf{Y}|\mathbf{Z},\mathbf{X})P(\mathbf{Z}|\mathbf{X}).
\end{eqnarray}
\shorteq

It is intractable to obtain an exact solution for maximizing Eq.~\ref{eq:split} due to the high time complexity. Inspired by the pre-ordering works in statistical machine translation, we propose a  \textbf{deterministic guiding decoding (DGD)}  strategy and a \textbf{non-deterministic guiding decoding (NDGD)} strategy to solve this problem.

The DGD strategy first generates the most probable pseudo-translation of the source sentence, and then translates the source sentence conditioned on it:
\shorteq
\begin{eqnarray}
    \mathbf{Z}^* &=& \mathop{\arg\max}_{\mathbf{Z}}P(\mathbf{Z}|\mathbf{X}), \\
    \mathbf{Y}^*  &=& \mathop{\arg\max}_{\mathbf{Y}} P(\mathbf{Y}|\mathbf{Z}^*,\mathbf{X}).\label{eq:split1}
    \shorteq
\end{eqnarray}

The DGD strategy is simple and effective, but the hard approximation may bring in some noises.

Different from the DGD strategy which utilizes a deterministic pseudo-translation, the NDGD strategy regards the probability distribution $\mathbf{\mathcal{Q}}$ of words to generate the most probable pseudo-translations as a latent variable, and models the translation as generating the target sentence according to   $\mathbf{\mathcal{Q}}$, i.e., Eq. \ref{eq:split1} is re-formulated as:
\shorteq
\begin{eqnarray}
\label{eq:ndgd}
    \mathbf{Y}^*
    &=& \mathop{\arg\max}_{\mathbf{Y}}P(\mathbf{Y}|\mathbf{\mathcal{Q}}, \mathbf{X}), 
   \shorteq
\end{eqnarray}
where $\mathbf{\mathcal{Q}}$ is defined as:
\shorteq
\begin{equation}
\label{eq:q}
    \mathbf{\mathcal{Q}}(z_i) = P(z_i|\mathbf{z}^*_{<i}, \mathbf{X}) = \frac{\exp\big(s(z_i)/T\big)}{\sum_{z_i'}\exp\big(s(z_i')/T\big)},
    \shorteq
\end{equation}
where $s(\cdot)$ is a score function of pseudo-translation (the input of softmax layer in the decoder) and $T$ is a temperature coefficient. 
Since $\mathbf{\mathcal{Q}}$ can be viewed as a non-deterministic form of the pseudo-translation, the translation with the NDGD strategy is also guided by the pseudo-translation.

To be specific, as shown in Figure \ref{fig:model}, the major difference between DGD and NDGD strategy is the inputs of decoder module (No. 2 dashed arrow), where the DGD strategy directly utilizes the word embeddings of generated pseudo-translation and the NDGD strategy utilizes the word embeddings weighted by the word probability of pseudo-translation. The detailed architecture of \ReorderNAT model is introduced in Appendix.

\subsection{Discussion}
In \ReorderNAT, the decoding space of generating pseudo-translation with reordering module is much smaller than that of the whole translation in NAT since the decoding vocabulary is limited to the words in the source sentence. The reordering module is more likely to be guided to one pseudo-translation among multiple alternatives. Therefore, \ReorderNAT could easily capture the reordering information compared to the original NAT by explicitly modeling with pseudo-translation as internal supervision. Besides, the candidates of the $i$-th word of the final translation can be narrowed to the translations of $z_i$ to some extent since $z_i$ is the $i$-th word in the pseudo-translation which indicates the corresponding source information of $y_i$. In other words, pseudo-translations could be viewed as a translation written in source language which helps the decoder to capture the dependencies among target words and choose words belonging to the same translation.

\subsection{Training}

In the training process, for each training sentence pair $(\mathbf{X}, \mathbf{Y}) \in D$, where $D$ is the training set, we first generate its corresponding pseudo-translation $\hat{\mathbf{Z}}$\footnote{The pseudo-translation construction details are in Appendix.}.
And then \ReorderNAT is optimized by maximizing a joint loss:
\shorteq
\begin{equation}
\mathcal{L} = \mathcal{L}_{R} + \mathcal{L}_{T},
\shorteq
\end{equation}
where $\mathcal{L}_{R}$ and $\mathcal{L}_{T}$ indicate the reordering and translation losses respectively. Formally, for both DGD and NDGD, the reordering loss $\mathcal{L}_{R}$ is defined as\footnote{Note that since $\mathbf{\mathcal{Q}}(\mathbf{Z}) = P(\mathbf{Z}|\mathbf{X})$, the reordering loss could also learn $\mathbf{\mathcal{Q}}$ for the NDGD approach.}
:
\shorteq
\begin{equation}
    \mathcal{L}_{R} = \sum_{(\mathbf{X}, \hat{\mathbf{Z}}, \mathbf{Y})\in D} \log P(\hat{\mathbf{Z}}|\mathbf{X}).
\shorteq
\end{equation}

For the DGD approach, the translation loss is defined as an overall maximum likelihood of translating pseudo-translation into the target sentence:
\shorteq
\begin{equation}
    \mathcal{L}_{T} = \sum_{(\mathbf{X}, \hat{\mathbf{Z}}, \mathbf{Y})\in D} \log P(\mathbf{Y}|\hat{\mathbf{Z}},\mathbf{X}),
\shorteq
\end{equation}

For the NDGD approach, the translation loss is defined as an overall maximum likelihood of decoding the target sentence from the conditional probability of pseudo-translation:
\shorteq
\begin{equation}
    \mathcal{L}_{T} = \sum_{(\mathbf{X}, \hat{\mathbf{Z}}, \mathbf{Y})\in D} \log P(\mathbf{Y}|\mathbf{\mathcal{Q}},\mathbf{X}).
\shorteq
\end{equation}

In particular, we use the trained model for the DGD approach to initialize the model for the NDGD approach since if $\mathbf{\mathcal{Q}}$ is not well trained, $\mathcal{L}_{T}$ will converge very slowly.

\section{Experiments}

\subsection{Datasets}

The main experiments are conducted on three widely-used machine translation tasks: WMT14 En-De ($4.5$M pairs), WMT16 En-Ro ($610$k pairs) and IWSLT16 En-De ($196$k pairs)\footnote{We use the prepossessed corpus provided by \citet{lee2018deterministic} at \url{https://github.com/nyu-dl/dl4mt-nonauto/tree/multigpu}.}. For WMT14 En-De task, we take newstest-2013 and newstest-2014 as validation and test sets respectively. For WMT16 En-Ro task, we employ newsdev-2016 and newstest-2016 as validation and test sets respectively. For IWSLT16 En-De task, we use test2013 for validation. 

We also conduct our experiments on Chinese-English translation which differs more in word order. The training set consists of $1.25$M sentence pairs extracted from the LDC corpora. We choose NIST 2002 (MT02) dataset as validation set, and NIST 2003 (MT03), 2004 (MT04), 2005 (MT05), 2006 (MT06) and 2008 (MT08) datasets as test sets. 


\subsection{Experimental Settings}
\label{sec:settings}

We follow most of the model hyperparameter settings in \citep{gu2018non,lee2018deterministic,wei2019imitation} for fair comparison. For IWSLT16 En-De, we use a $5$-layer Transformer model ($d_{model}=278$, $d_{hidden}=507$, $n_{head}=2$, $p_{dropout}=0.1$) and anneal the learning rate linearly (from $3 \times 10^{-4}$ to $10^{-5}$) as in \citep{lee2018deterministic}. For WMT14 En-De, WMT16 En-Ro and Chinese-English translation, we use a $6$-layer Transformer model ($d_{model}=512$, $d_{hidden}=512$, $n_{head}=8$, $p_{dropout}=0.1$) and adopt the warm-up learning rate schedule \citep{vaswani2017attention} with $t_{warmup}=4000$. For the GRU reordering module, we set it to have the same hidden size with the Transformer model in each dataset. We employ label smoothing of value $\epsilon_{ls}=0.15$ and  utilize the sequence-level knowledge distillation~\citep{kim2016sequence} for all tasks. 
For each dataset, we also select the optimal guiding decoding strategy according to the model performance on validation sets\footnote{The performance of our model using DGD and NDGD on all tasks are shown in Appendix.}.
We also set $T$ in Eq.~\ref{eq:q} to $0.2$ according to a grid search on the validation set (see more details in Appendix).
We measure the model inference speedup on the validation set of IWSLT16 En-De task with a NVIDIA P40 GPU and set batch size to $1$. 

\subsection{Baselines}
In the experiments, we compare \ReorderNAT (NAT) and \ReorderNAT (AT) which utilize an NAT and an AT reordering modules respectively with several baselines\footnote{We provide the results of autoregressive models using greedy search and more non-autoregressive baselines in Appendix.}.

We select three models as our autoregressive baselines:
(1) \textbf{Transformer$_{full}$}~\citep{vaswani2017attention}, which is the teacher model used in the knowledge distillation and of which the hyperparameters are described in experimental settings.
(2) \textbf{Transformer$_{one}$}, a lighter version of Transformer, of which the decoder layer number is $1$.
(3) \textbf{Transformer$_{gru}$}, which replaces the decoder of Transformer$_{full}$ with a single-layer GRU~\citep{cho2014learning}. 
And we set the beam size to $4$ in the experiments.

Besides, we compare with several typical NAT models, which also have the ability  to alleviate the multimodality problem and are highly relevant to our work:
(1) \textbf{NAT-IR}~\citep{lee2018deterministic}, which adopts an iterative refinement mechanism for better translation quality;
(2) \textbf{NAT-FS}~\citep{shao2019retrieving}, which introduces the autoregressive property to the top decoder layer of NAT;
(3) \textbf{FlowSeq-base}~\citep{ma2019flowseq}, which uses generative flow to help model dependencies within target sentences. For fair comparison, We use the ``base'' version as it has comparable model size with our model;
(4) \textbf{imitate-NAT}~\citep{wei2019imitation}, which imitates the behavior of an AT model.
(5) \textbf{CMLM-small}~\citep{ghazvininejad2019mask}, which is built on a conditional masked language model and also iteratively refines the translation. We use the ``small'' version for fair comparison;
(6) \textbf{NART-DCRF}~\citep{sun2019fast}, which uses CRF to capture the word dependencies;
(7) \textbf{LevT}~\citep{gu2019levenshtein}, which models the sequence generation as multi-step insertion and deletion operations.



%

\begin{table*}[!t]
  \centering
  \small
  {
  \setlength{\tabcolsep}{3pt}
  \begin{tabular}{llccccccr}
    \toprule
    \multicolumn{1}{l}{\multirow{2}{*}{Model}} & \multicolumn{1}{l}{\multirow{2}{*}{Decoder Architecture}} & \multicolumn{1}{l}{\multirow{2}{*}{Multi-Step}}     & \multicolumn{2}{c}{WMT14}     & \multicolumn{2}{c}{WMT16}     & IWSLT16       & \multirow{2}{*}{Speedup}\\
                                                & &                               & En$\to$De     & De$\to$En     & En$\to$Ro     & Ro$\to$En     & En$\to$De     &  \\
    \midrule
    \bf{Autoregressive Models}\\
    Transformer$_{full}$                        & AT-TM$\times N$ & -               & 27.17         & 31.95	        & 32.86         & 32.60          & 31.18         & 1.00$\times$\\
    Transformer$_{one}$                         & AT-TM$\times 1$ & -               & 25.52         & 29.31         & 30.61         & 31.23         & 29.52         & 2.42$\times$ \\
    Transformer$_{gru}$                         & AT-GRU$\times 1$ & -               & 26.27         & 30.62         & 30.41         & 31.23         & 29.26         & 3.10$\times$ \\
    \midrule
    \multicolumn{3}{l}{\bf{Non-Autoregressive Models}}\\
    NAT-IR (iter=1)                             & NAT-TM$\times N$ & -              & 13.91         & 16.77         & 24.45         & 25.73         & 22.20         & 8.98$\times$\\
    NAT-IR (iter=10)                            & NAT-TM$\times N$ & $\surd$              & 21.61         & 25.48         & 29.32         & 30.19         & 27.11         & 1.55$\times$\\
    NAT-FS                      & NAT-TM$\times N$-$1$+AT-TM$\times 1$ & -           & 22.27         & 27.25         & 30.57         & 30.83         & 27.78         & 3.38$\times$\\
    FlowSeq-base           & FlowStep & -  & 21.45         & 26.16         & 29.34         & 30.44         & -             & $<$1.5$\times$\\
    FlowSeq-base+NPD (s=30)  & FlowStep & - & 23.48         & 28.40         & \underline{31.75}         & \underline{32.49}         & -             & $<$1.5$\times$\\
    imitate-NAT	                                & NAT-TM-imitate$\times N$ & -      & 22.44	        & 25.67	        & 28.61	        & 28.90         & 28.41	        & 18.6$\times$\\
    imitate-NAT+LPD (s=7)	                            & NAT-TM-imitate$\times N$ & -       & 24.15         & 27.28	        & 31.45	        & 31.81         & \textbf{30.68}	        & 9.70$\times$\\
    CMLM-small (iter=10) & NAT-TM$\times$ N & $\surd$ & 25.51         & 29.47         & 31.65         & \underline{32.27}         & -             & $<$1.5$\times$\\
    NART-DCRF            & NAT-TM$\times$ N+DCRF$\times$ 1& -  & 23.44           & 27.22         & -             & -             & -             & 10.4$\times$\\
    NART-DCRF+LPD (s=19)   & NAT-TM$\times$ N+DCRF$\times$ 1& -   & \underline{26.80}       & 30.04         & -             & -             & -             & 4.39$\times$\\
    LevT                  & NAT-TM$\times$ N     & $\surd$ & \underline{27.27}       & -         & -      & \underline{33.26}      & -             & 4.01$\times$\\
    \midrule
    \bf{Our Models}\\
    \ReorderNAT (NAT)                             & NAT-TM$\times 1$ & -               &  22.79	      & 27.28          & 29.30         &   29.50       & 25.29       & 16.11$\times$\\
    \ReorderNAT (NAT)+LPD (s=7)                     & NAT-TM$\times 1$ & -              & 24.74    & 29.11	& 31.16     & 31.44   &  27.40   & 7.40$\times$\\
    \ReorderNAT (AT)            & NAT-TM$\times N$-$1$ & -          & \textbf{26.49}	              & \textbf{31.13}              & \textbf{31.70}         & \textbf{31.99}               & 30.26 & 5.96$\times$\\
    \bottomrule
  \end{tabular}}
  \caption{Overall results of AT and NAT models in BLEU score on the test sets of WMT14 and WMT16, and validation set of IWSLT16. 
  ``AT-TM'' and ``NAT-TM'' denote the AT Transformer and NAT Transformer decoder block respectively. ``AT-GRU'' denotes the AT GRU decoder block. ``NAT-TM-imitate'' denotes the NAT Transformer decoder block with the imitation module. ``FlowStep'' denotes flow steps~\citep{ma2019flowseq}. ``DCRF'' denotes a CRF layer with dynamic transition~\citep{sun2019fast}. ``NPD'' denotes noisy parallel decoding~\citep{gu2018non}, ``LPD'' denotes length parallel decoding~\citep{wei2019imitation}, and ``s'' denotes sample size. ``iter'' denotes translation refinement iterations. Better BLEU scores with {\em low speedup} are underlined.}
  \label{tab:overall-results}
	\shrink
\end{table*}



\subsection{Overall Results}

We compare \ReorderNAT (NAT) and \ReorderNAT (AT) that utilize an NAT reordering module and an AT reordering module respectively with all baseline models.
All the results are shown in Table~\ref{tab:overall-results}. From the table, we can find that: 

(1) 
Excluding six better BLEU scores with significant low speedup, \ReorderNAT (AT) achieves the best performance on most of the benchmark datasets, which is even close to the AT model with smaller than $1$ BLEU gap ($26.49$ vs. $27.17$ on WMT14 En$\to$De task, $31.99$ vs. $32.60$ on WMT16 Ro$\to$En task, $30.26$ vs. $31.18$ on IWSLT16 En$\to$De task). It is also worth mentioning that although \ReorderNAT utilizes a small AT model to better capture reordering information, it could still maintain low translation latency (about $16\times$ speedup for \ReorderNAT (NAT) and $6\times$ speedup for \ReorderNAT (AT)). Compared to Transformer$_{one}$ and Transformer$_{gru}$, \ReorderNAT (AT) uses a much smaller vocabulary in the AT reordering module, which is limited to the words in the source sentence and makes it faster. Besides, unlike NAT-IR, CMLM-small and LevT, our model can decode all target words in parallel without multiple iterations, which helps maintain the efficiency.


(2) \ReorderNAT (NAT) and \ReorderNAT (NAT)+LPD also gain significant improvements compared to most existing  NAT models. 
It verifies the reordering information modeled by \ReorderNAT could effectively guide the translation.


(3) A small AT model with close latency to large NAT models could perform much better in modeling reordering information\footnote{The decoding speed of \ReorderNAT (AT) is limited by the autoregressive property of the reordering module, which is the main drawback of our model. How to further improve its speed is a future direction we would like to pursue.}. On all benchmark datasets, \ReorderNAT (AT) with small AT GRU reordering module achieves much better translation quality than that with large NAT model ($2$-$5$ BLEU scores). Moreover, we find that the AT model Transformer$_{one}$ and Transformer$_{gru}$ with a single-layer AT Transformer or GRU for decoding could also outperform most existing NAT models while maintaining acceptable latency ($2.42\times$ and $3.10\times$ speedup respectively). The reason is that a major potential performance degradation of NAT models compared to AT models comes from the difficulty of modeling the
reordering information, which is neglected by most existing NAT models but can be well modeled by the small AT module\footnote{We conduct experiments and find that our model outperforms the AT model by a big margin when replacing the predicted pseudo-translation with the ground-truth ones. This also indicates the main multimodality problem on NAT comes from the difficulty of modeling the reordering information.}.


\subsection{Results on Chinese-English Translation}

\begin{table}[!t]
  \centering
  \small
  {
  \begin{tabular}{lcccc}
    \toprule
    Model                   & MT02*      & MT03       & MT04   & MT05   \\
    \midrule
    \multicolumn{3}{l}{\bf{Autoregressive Models}}\\
    Transformer$_{full}$    & 46.11	    & 43.74	    & 45.59	& 44.11 \\
    Transformer$_{one}$     & 43.60	    & 41.24	    & 43.39	& 41.62 \\
    Transformer$_{gru}$     & 43.68	    & 40.55	    & 43.02	& 40.73 \\
    \midrule
    \multicolumn{3}{l}{\bf{Non-Autoregressive Models}}\\
    imitate-NAT             & 33.77	    & 32.29	    & 34.83	& 31.96       \\
    \ReorderNAT (NAT)         & 37.99	    & 36.03	    & 38.17 & 36.07 \\
    \ReorderNAT (AT)          & \textbf{45.22}	    & \textbf{43.20}	    & \textbf{44.89}	& \textbf{43.45} \\
    \bottomrule
    \end{tabular}}
    \caption{BLEU scores on Chinese-English translation. * indicates the validation set. More results are in Appendix.}
    \label{tab:chinese_english}
    \shrink
\end{table}

\begin{figure}[!t]
     \centering
     \includegraphics[width=0.70\columnwidth]{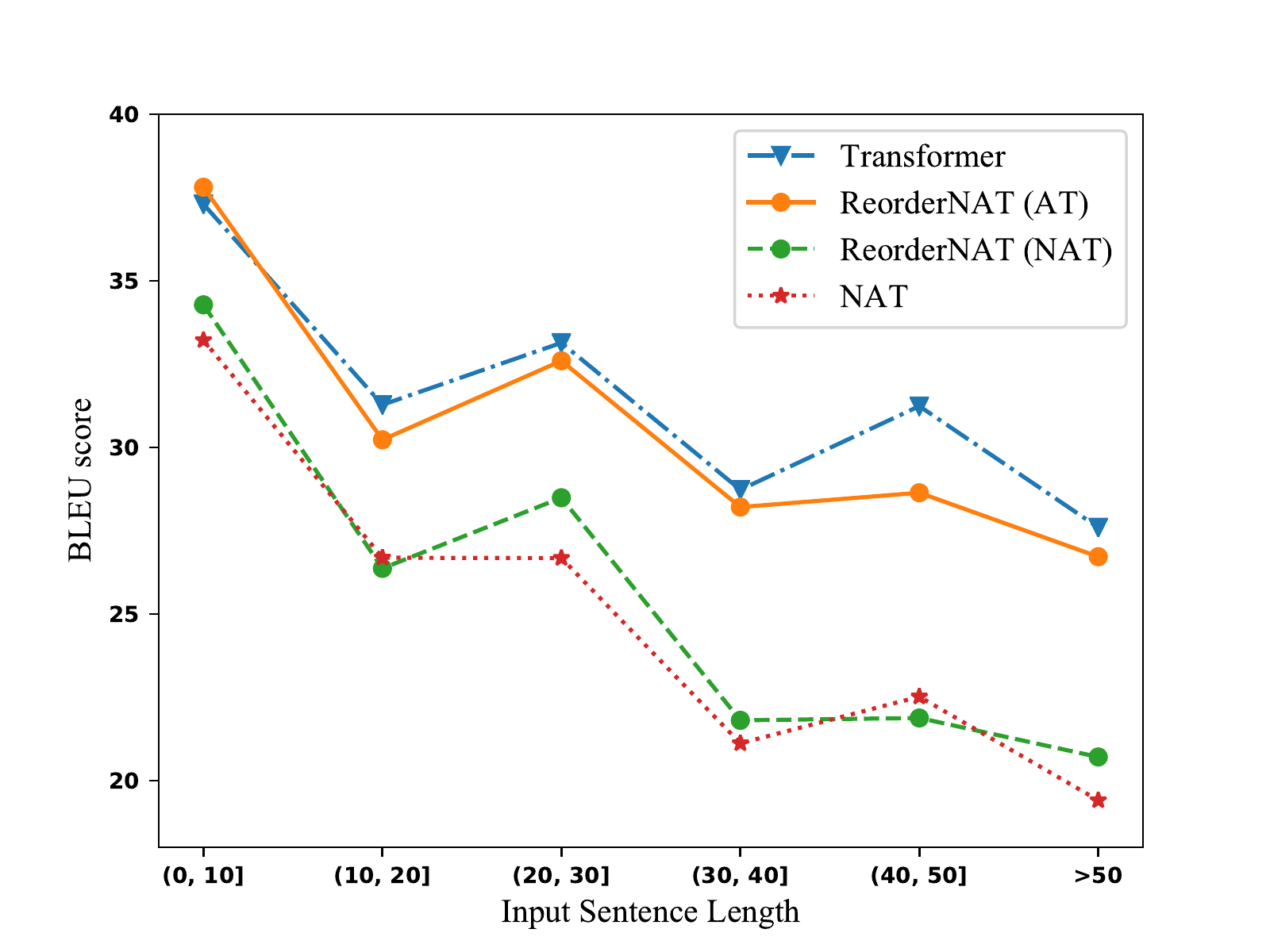}
     \caption{Translation quality on the IWSLT16 validation set over various input sentence lengths.}
     \label{fig:length} 
     \shrink
 \end{figure}

To show the effectiveness of modeling reordering information in NAT, we compare \ReorderNAT with baselines on Chinese-English translation since the word order between Chinese and English is more different than that between German and English (En-De). From Table~\ref{tab:chinese_english}, we can find that in Chinese-English translation, \ReorderNAT (AT) achieves much more improvements ($6$-$7$ BLEU points) compared to \ReorderNAT (NAT) and imitate-NAT. 
The reason is that more different word order in Chinese-English translation makes the decoding search space more complicated, which could be effectively alleviated by \ReorderNAT.

\subsection{Translation Quality v.s. Sentence Lengths}

Figure~\ref{fig:length} shows the BLEU scores of translations generated by the Transformer (AT model), the NAT model (\ReorderNAT without the reordering module), \ReorderNAT (NAT) and \ReorderNAT (AT) on the IWSLT16 validation set with respect to input sentence lengths\footnote{Results on WMT14 En-De and WMT16 En-Ro tasks are in Appendix.}. 
We can observe that: 

(1) \ReorderNAT (NAT) and \ReorderNAT (AT) achieve significant improvement compared to the NAT model for most lengths and \ReorderNAT (AT) achieves nearly comparable performance to Transformer. It verifies the reordering information modeled by \ReorderNAT could effectively help word selection and improve the translation quality.

(2) \ReorderNAT (AT) achieves much better translation performance than the NAT model for sentences longer than $20$ words, of which word order tends to be more different. 
Together with the results on Chinese-English translation (Table~\ref{tab:chinese_english}), we can conclude that NAT is weak on word reordering and our model is more effective especially when word order is more different.


\subsection{Multimodality Related Error Reduction}
In this section, we investigate how our reordering module reduces the multimodality errors in NAT. Specially, we evaluate the RIBES~\citep{isozaki2010automatic} score and the reduction of duplicate and missing words (two most typical multimodality related errors). The results are shown in Table~\ref{tab:multimodality-error}, where ``Dup'' and ``Mis'' denote the relative increment of duplicate and missing token ratios compared with the Transformer$_{full}$ model respectively\footnote{The formal definition of metrics ``Dup/Mis'' can be found in \citet{ran2020learning} and the results on WMT14 En-De and WMT16 En-Ro tasks are in Appendix.}
, and NAT is \ReorderNAT without the reordering module. From the table, we can observe that: 

(1) Our three \ReorderNAT models achieve higher RIBES scores than the NAT model, validating our reordering module can help capture the word order difference between source and target language. Moreover, the \ReorderNAT (AT) model performs the best in RIBES, indicating the AT reordering module can model reordering information more effectively than the NAT reordering module.

(2) Compared with the NAT model, both Dup and Mis are significantly better for the three \ReorderNAT models, indicating \ReorderNAT is effective for alleviating the multimodality problem.

\begin{table}[!t]
  \centering
  \small
  \setlength{\tabcolsep}{2.4pt}
  {
  \begin{tabular}{lrrrr}
    \toprule
     Model & BLEU   & RIBES   &  Dup $\downarrow$  &   Mis $\downarrow$\\
    \midrule
    Transformer$_{full}$        &   31.18   & 83.74  & \multicolumn{1}{c}{-} & \multicolumn{1}{c}{-} \\
    \midrule
    NAT                         &   24.57   & 82.21   & 50.09  & 9.09  \\
    \midrule
    \ReorderNAT (NAT)           &   25.29   & 82.35   & 37.52  &  7.35 \\
    \ReorderNAT (NAT)+LPD (s=7) &   27.04   & 83.21   & 24.31  &  5.59 \\
    \ReorderNAT (AT)            &   30.26   & 83.55   & 2.84  &    0.52 \\
    \bottomrule
    \end{tabular}}
    \caption{Relative increment of duplicate (``Dup'') and missing (``Mis'') token ratios on the IWSLT16 validation set. Smaller is better.}
    \label{tab:multimodality-error}
    \shrink
\end{table}

\subsection{Case Study}

\begin{table*}
	\small
	\centering
    {
	\begin{tabular}{p{0.085\textwidth} p{0.065\textwidth} p{0.765\textwidth}}
		\toprule
		Source  &  &eventually , after a period of six months of brutal war and a toll rate of almost 50,000 dead , we managed to liber\_ate our country and to t\_opp\_le the ty\_rant \\
		\midrule
		Reference   & &schließlich , nach einem Zeitraum von sechs Monaten bru\_talen Krieges und fast 50.000 Toten , gelang es uns , unser Land zu befreien und den Tyran\_nen zu stürzen .\\
		\toprule
		NAT & Translation &schließlich , nach einer \tl{\ \ } von sechs Monaten bru\_\hl{bru\_}\tl{\ \ } Krieg und einer Z\_rate fast 50.000 \hl{50.000} \tl{\ \ } , schafften wir es \hl{geschafft} , unser Land \tl{\ \ } befreien \hl{befreien} und den Ty\_r\_ann \tl{\ \ } \hl{ann entgegen\_entgegen\_deln} .\\
		\midrule
		 \ReorderNAT (NAT)  & Pseudo-Translation &  eventually , after a period of six \tl{\ \ } brutal brutal war and a toll of almost 50,000 dead , managed we managed \hl{managed} , \tl{\ \ } \rl{country} country to liber\_\tl{\ \ } and the \hl{ty\_ty\_rant rant opp\_opp\_opp\_}.\\
		 \cmidrule{2-3}
		 & Translation &schließlich , nach einer Zeit von sechs \tl{\ \ } bru\_talen Krieges und einer Z\_von fast 50.000 Toten , schafften wir es \hl{geschafft} , unser Land zu befreien und den \hl{Ty\_r\_r\_ten zu\_zu\_ieren} .\\
		\midrule
		 \ReorderNAT (AT)   & Pseudo-Translation & eventually , after a period of six months brutal brutal war and a toll toll rate of almost 50,000 dead , managed we \rl{managed} , our country to liber\_\tl{\ \ } and the ty\_ty\_rant to \rl{liber\_}.\\ 
         \cmidrule{2-3}
		 & Translation & schließlich , nach einer Zeitraum von sechs Monaten bru\_talen Krieg und einer Z\_oll\_rate von fast 50.000 Toten , schafften wir es , unser Land zu befreien und den Ty\_r\_ann zu \rl{reparieren} .\\
		\bottomrule
	\end{tabular}}
	\caption{Translation examples of NAT baseline and \ReorderNAT. We color the missing, duplicate and wrong words in \tl{red}, \hl{yellow} and \rl{green} respectively. And we use \_ to concatenate sub-words.}
	\label{tab:case_study}
	\shrink
\end{table*}

Table~\ref{tab:case_study} shows example translations of the NAT model (\ReorderNAT without the reordering module), \ReorderNAT (NAT) and \ReorderNAT (AT). We find that the problem of missing and duplicate words are severe in the translation (both $5$ occurrences) of the NAT model, while this problem is effectively alleviated by \ReorderNAT. Moreover, we find that most of the missing, duplicate or wrong words in the translation of \ReorderNAT (NAT) and \ReorderNAT (AT) come from the errors in the pseudo-translation, demonstrating that NAT models could well translate the pseudo-translation which is in the the target language word order, and the remaining problem of NAT lies on modeling reordering information.

\section{Related Work}

\subsection{Non-Autoregressive Neural Machine Translation}

\citet{gu2018non} first proposed the non-autoregressive neural machine translation (NAT), which enables parallel decoding for neural machine translation (NMT) and significantly accelerates the inference of NMT. However, its  performance degrades greatly since it discards the sequential dependencies among words in the target sentence. Recently, a variety of works have been investigated to improve the performance of NAT in various aspects including~\citep{guo2019non,bao2019pnat} which enhance the representation of decoder with source information; 
\citep{libovicky2018end,shao2019retrieving,shao2019minimizing,ghazvininejad2020aligned} which optimize models with respect to sequence-level loss functions;
\citep{zhou2020improving} which incorporates monolingual data for training;
\citep{wang2019non,li2020lava} which alleviate the multimodality problem by making the decoder representation more informative via regularization or speical decoding strategies; 
\citep{li2019hint,wei2019imitation,guo2019fine,liu2020task,sun2020an} which guide the learning of NAT models with AT models;
\citep{kaiser2018fast,akoury2019syntactically,lee2020iterative} which introduce latent variables to guide the decoding process of NAT models;
and \citep{wang2018semi,lee2018deterministic,gu2019levenshtein,stern2019insertion,ghazvininejad2019mask,ghazvininejad2020semi,kasai2020non,tu2020engine,guo2020jointly,ran2020learning} which extend one-step NAT to multi-step NAT and generate translations iteratively.
Different from existing works, we propose to explicitly model reordering information in NAT models, which serves as a proxy in capturing the dependencies of the target words and encourages the decoder to choose words belonging to the same translation to alleviate the multimodality problem. This work intends to enhance the translation quality of one-step NAT models and has the potential to improve the performance of each iteration of the multi-step NAT methods without loss of efficiency.

\subsection{Modelling Reordering Information in Machine Translation}

Re-ordering model is a key component in statistical machine translation (SMT), which handles the word order differences between source and target language. There has been a large number
of works focusing on word reordering in SMT, including deterministic reordering approaches~\citep{xia2004improving,collins2005clause,wang2007chinese,li2007probabilistic}, which find a single optimal reordering of the source sentence; non-deterministic reordering approaches~\citep{kanthak2005novel,zhang2007chunk} which encode multiple alternative reorderings into a word lattice and remain the choosing strategy of best path in the decoder; and target word reordering approaches~\citep{bangalore2007statistical} which first select target lexicals and then reorder them to form final sentence.
In neural machine translation (NMT), it has been shown that the attention mechanism~\citep{bahdanau2014neural} could implicitly capture the reordering information to some extent. \citet{zhang2017incorporating} presented three distortion models to further incorporate reordering knowledge into attention-based NMT models. \citet{chen2019neural} introduced a reordering mechanism for NMT models to learn the reordering embedding of a word based on its contextual information.  Except for incorporating reordering knowledge in attention mechanism, researchers also proposed to learn to reorder the source-side word orders according to the sentence structure in target language with neural networks~\citep{du2017pre,kawara2018recursive,zhao2018exploiting}.
This work empirically justifies that reordering information is essential for NAT.

\section{Conclusion and Future Work}


In this work, to address the multimodality problem in NAT, we propose a novel NAT framework named \ReorderNAT which explicitly models the reordering information in the decoding procedure. We further introduce deterministic and non-deterministic guiding decoding strategies to utilize the reordering information to encourage the decoder to choose words belonging to the same translation. Experimental results on several widely-used benchmarks show that our \ReorderNAT model achieves better performance than most existing NAT models, and even achieves comparable translation quality as AT model with a significant speedup. 
We believe that to well model the reordering information is a potential way towards better NAT.

\bibliography{emnlp2020}
\bibstyle{aaai21}

\appendix

\section{Model Architecture}

\begin{figure*}
    \centering
    \includegraphics[width=1.0\textwidth]{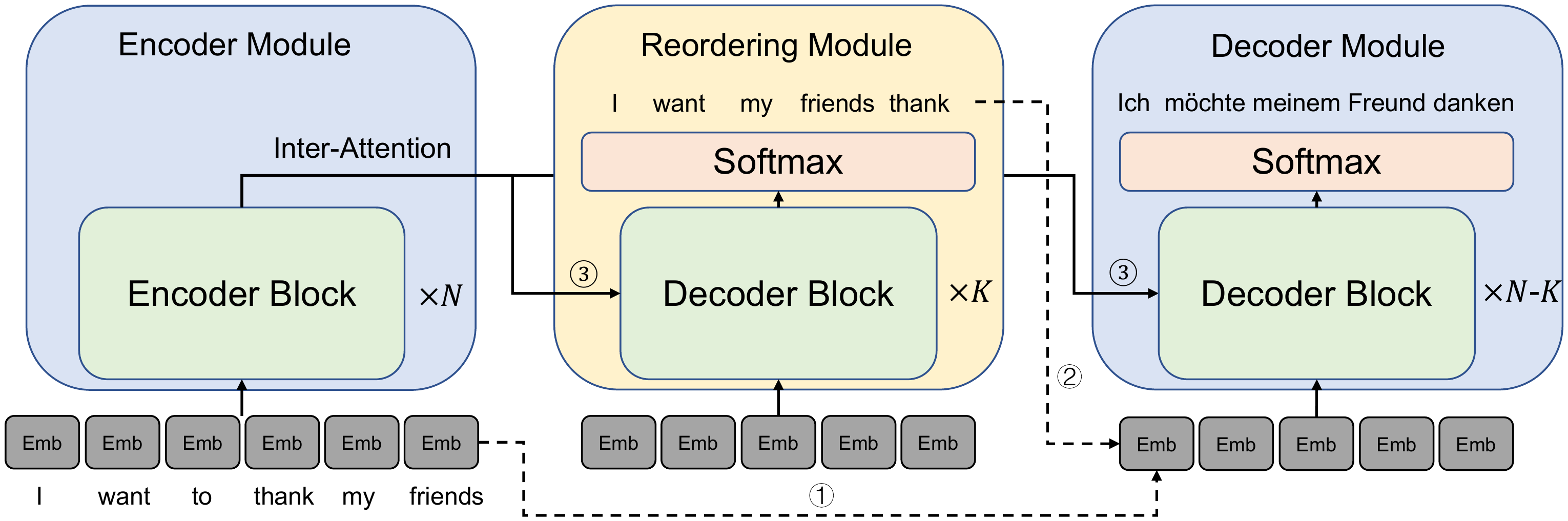}
    \caption{{ The framework of our \ReorderNAT model}.}
    \label{fig:model_detail} 
\end{figure*}

In this section, we will introduce the details of the encoder, reordering and decoder modules in Figure~\ref{fig:model_detail} respectively.

\subsection{Encoder and Decoder Blocks}

We first introduce the encoder and decoder blocks used in the three modules.

In our proposed \ReorderNAT model,  we employ the \textbf{Transformer encoder block}~\citep{vaswani2017attention} as our encoder block, which is composed by a multi-head self-attention layer and two linear transformation layers. It could be formulated as:
\begin{equation}
    E(\mathbf{H}) = \texttt{FFN}(\texttt{Self-Att}(\mathbf{H})),
    \label{enc-block}
\end{equation}
where $\mathbf{H}$ is the input embeddings of the encoder block, $\texttt{FFN}(\cdot)$ denotes the two linear transformation layers and  $\texttt{Self-Att}(\cdot)$ is the multi-head self-attention layer.

We employ two types of decoder blocks:

(1) \textbf{Transformer decoder block}~\citep{vaswani2017attention}, which is composed by a multi-head self-attention layer, a multi-head inter-attention layer and two linear transformation layers. It is formulated as:
\shorteq
\begin{eqnarray}
    D(\mathbf{H}, \mathbf{S}) &=& \texttt{FFN}(\texttt{Inter-Att}(\mathbf{S},\nonumber\\
    &&(\texttt{Self-Att}(\mathbf{H}))),
\shorteq
\end{eqnarray}
where $\mathbf{H}$ is the input embeddings of the decoder block, $\mathbf{S}$ is the hidden representation of the source sentence, and $\texttt{Inter-Att}(\cdot)$ is the multi-head inter-attention layer. 

(2) \textbf{GRU decoder block}, which consists of a multi-head inter-attention layer and a GRU layer. Its hidden state is computed as
\begin{equation}
    \mathbf{D}_i=\texttt{GRU}(\mathbf{D}_{i-1}, [\bm{C}_{i-1}; \mathbf{H}_{i-1}]),
\end{equation}
where $\mathbf{H}_{i-1}$ is the embedding of the previous word in the target sentence, $\bm{C}_{i-1}$ is calculated by the inter-attention to the encoder representation, and $\texttt{GRU}(\cdot)$ is a gated recurrent unit (GRU)~\citep{cho2014learning}.

\subsection{Encoder Module}
In the \ReorderNAT model, we feed the input sentence embedding $\texttt{Emb}(\mathbf{X})$ into an encoder $E^N(\cdot)$ consisting of $N$ encoder blocks $E(\cdot)$ (see Eq.~\ref{enc-block}) to obtain the hidden representation $\mathbf{S}$ of the source sentence $\mathbf{X}$:
\begin{equation}
    \mathbf{S} = E^N(\texttt{Emb}(\mathbf{X})).
\end{equation}

\subsection{Reordering Module}

In the reordering module, we employ an NAT model with Transformer decoder block or an AT model with GRU decoder block as the  pseudo-translation decoder. Formally, for the \textbf{NAT Reordering Module}, we utilize the uniform copied embeddings as~\citep{gu2018non} for the input of the decoder, and then obtain the hidden representation $\mathbf{R}$ in reordering module:
\begin{equation}
    \mathbf{R} = D(\texttt{Uniform-Copy}(\texttt{Emb}(\mathbf{X})), \mathbf{S}).
\end{equation}

After obtaining the decoder hidden state $\mathbf{R}$, the conditional probability $P(z_i|\mathbf{X})$ of the $i$-th word in the pseudo-translation is computed as:
\begin{equation}
\label{eq:z}
    P(z_i|\mathbf{X}) = \texttt{Softmax}(\mathbf{R}_i),
\end{equation}
where $\mathbf{R}_i$ is the $i$-th column of $\mathbf{R}$.

For the \textbf{AT Reordering Module}, we use a greedy search algorithm to decode the pseudo-translation word-by-word. When decoding the $i$-th word $z_i$ in the pseudo-translation, we obtain its hidden representation as: 
\begin{equation}
    \mathbf{R}_i=\texttt{GRU}(\bm{R}_{i-1}, [\bm{C}_{i-1}; Emb(z^*_{i-1})]),
\end{equation}
where $z*_{i-1}$ is the previous word in the decoded pseudo-translation and $\bm{C}_{i-1}$ is calculated by the inter-attention to the encoder representation. After obtaining the hidden representation, the conditional probability $P(z_i|\mathbf{X})$ is also calculated as Eq. \ref{eq:z}.

\subsection{Decoder Module}

In the decoder module, we employ an NAT decoder with $N$-$1$ layers of Transformer decoder block. We utilize two types of input representation for deterministic guiding decoding (DGD) strategy and non-deterministic guiding decoding (NDGD) strategy. Formally, for \textbf{DGD strategy}, we use the word embeddings of the predicted pseudo-translation $\mathbf{Z}^*$ as the input of the decoder, and then obtain the final translation probability:
\begin{equation}
    P(\mathbf{Y}|\mathbf{Z}^*, \mathbf{X}) = \texttt{Softmax}( D^{N-1}(\texttt{Emb}(\mathbf{Z}^*), \mathbf{S})),
\end{equation}
where $D^{N-1}(\cdot)$ indicates stacking $N-1$ layers of Transformer decoder block $D(\cdot)$.

For \textbf{NDGD strategy}, we use the  word embeddings weighted by the conditional probability $\mathbf{\mathcal{Q}}$ (i.e. $\mathbf{\mathcal{Q}}(\mathbf{Z}) = P(\mathbf{Z}|\mathbf{X})$) of pseudo-translation as the input of the decoder, and then obtain the final translation probability:
\begin{equation}
    P(\mathbf{Y}|\mathbf{\mathcal{Q}}, \mathbf{X}) = \texttt{Softmax}( D^{N-1}(\mathbf{\mathcal{Q}}^T\texttt{Emb}(\mathbf{X}), \mathbf{S})).
\end{equation}

\section{Pseudo-Translation Construction}

We use the fast\_align tool\footnote{\url{https://github.com/clab/fast_align}} to generate the pseudo-translation used for training in our experiments. Specifically, we use fast\_align to obtain word alignments between source and target sentences\footnote{Both the target sentences from the original training data and those produced by the teacher model are used in sequence-level knowledge distillation~\citep{kim2016sequence} are used.}. The maximum number of source words aligned to each target word is set to 1. For each target word, the corresponding word in the pseudo-translation is set as the source word aligned to it (We add a special symbol if the target word does not align to any source word.).

\section{Other Experimental Results}
\subsection{Overall Performance}
In Table~\ref{tab:overall-results:more}, we compare our \ReorderNAT model with more NAT baselines: NAT-FT~\citep{gu2018non}, NAT-IR~\citep{lee2018deterministic}, NAT-FS~\citep{shao2019retrieving}, FlowSeq-base~\citep{ma2019flowseq}, imitate-NAT~\citep{wei2019imitation}, NAT-REG~\citep{wang2019non}, LV NAR~\citep{shu2019latent}, NART~\citep{li2019hint}, TCL-NAT~\citep{liu2020task}, CMLM-small~\citep{ghazvininejad2019mask}, NAT-EM~\citep{sun2020an}, FCL-NAT~\citep{guo2019fine}, NART-DCRF~\citep{sun2019fast} and LevT~\citep{gu2019levenshtein}.

The performance on the validation sets are shown in Table~\ref{tab:overall-results:dev} \footnote{Note that the results on the IWSLT16 dataset shown in Table~\ref{tab:overall-results:more} are on its validation set as its test set corpus is not available.}.

\begin{table*}
  \centering
  \small
  \resizebox{1.0\textwidth}{!}{
  \setlength{\tabcolsep}{3pt}
  \begin{tabular}{llccccccr}
    \toprule
    \multicolumn{1}{l}{\multirow{2}{*}{Model}} & \multicolumn{1}{l}{\multirow{2}{*}{Decoder Architecture}} & \multicolumn{1}{l}{\multirow{2}{*}{Multi-Step}}     & \multicolumn{2}{c}{WMT14}     & \multicolumn{2}{c}{WMT16}     & IWSLT16       & \multirow{2}{*}{Speedup}\\
                                                & &                               & En$\to$De     & De$\to$En     & En$\to$Ro     & Ro$\to$En     & En$\to$De     &\\
    \midrule
    \bf{Autoregressive Models}\\
    Transformer$_{full}$ (b=1)                  & AT-TM$\times N$ & -              & 26.52         & 31.13         & 31.79         & 31.99          & 30.47         & 1.29$\times$\\
    Transformer$_{full}$ (b=4)                  & AT-TM$\times N$ & -              & 27.17         & 31.95	        & 32.86         & 32.60          & 31.18         & 1.00$\times$\\
    Transformer$_{one}$ (b=1)                   & AT-TM$\times 1$ & -              & 25.03         & 28.49         & 30.40         & 30.75         & 28.70          & 3.53$\times$\\
    Transformer$_{one}$ (b=4)                   & AT-TM$\times 1$ & -              & 25.52         & 29.31         & 30.61         & 31.23         & 29.52         & 2.42$\times$\\
    Transformer$_{gru}$ (b=1)                   & AT-GRU$\times 1$ & -             & 25.36         & 29.44         & 29.70         & 30.30         & 28.22         & 4.97$\times$\\
    Transformer$_{gru}$ (b=4)                   & AT-GRU$\times 1$ & -            & 26.27         & 30.62         & 30.41         & 31.23         & 29.26         & 3.10$\times$\\
    \midrule
    \multicolumn{3}{l}{\bf{Non-Autoregressive Models}}\\
    NAT-FT                                      & NAT-TM$\times N$ & -             & 17.69         & 21.47         & 27.29         & 29.06         & 26.52         & 15.6$\times$\\
    NAT-FT+NPD (s=100)                          & NAT-TM$\times N$ & -             & 19.17         & 23.20         & 29.79         & 31.44         & 28.16         & 2.36$\times$\\
    NAT-IR (iter=1)                             & NAT-TM$\times N$  & -             & 13.91         & 16.77         & 24.45         & 25.73         & 22.20         & 8.98$\times$\\
    NAT-IR (iter=10)                            & NAT-TM$\times N$  & $\surd$             & 21.61         & 25.48         & 29.32         & 30.19         & 27.11         & 1.55$\times$\\
    NAT-FS                      & NAT-TM$\times N$-$1$+AT-TM$\times 1$ & -         & 22.27         & 27.25         & 30.57         & 30.83         & 27.78         & 3.38$\times$\\
    FlowSeq-base           & FlowStep  & - & 21.45         & 26.16         & 29.34         & 30.44         & -             & $<$1.5$\times$\\
    FlowSeq-base+NPD (s=30)  & FlowStep & -  & 23.48         & 28.40         & \underline{31.75}         & \underline{32.49}         & -             & $<$1.5$\times$\\
    imitate-NAT	                                & NAT-TM-imitate$\times N$  & -     & 22.44	        & 25.67	        & 28.61	        & 28.90         & 28.41	        & 18.6$\times$\\
    imitate-NAT+LPD (s=7)	                            & NAT-TM-imitate$\times N$ & -     & 24.15         & 27.28	        & 31.45	        & 31.81         & \textbf{30.68}	        & 9.70$\times$\\
    NAT-REG                                     & NAT-TM$\times N$ & -             & 24.61         & 28.90         & -             & -             & 27.02         & 15.1$\times$\\
    LV NAR                 & NAT-TM$\times$ N & - & 25.10         & -             & -             & -             & -             & 6.8$\times$\\
    NART     & NAT-TM$\times$ N & -  & 21.11         & 25.24         & -             & -             & -             & 30.2$\times$\\
    NART+LPD (s=9)     & NAT-TM$\times$ N & -  & 25.20         & 29.52         & -             & -             & -             & 17.8$\times$\\
    TCL-NAT              & NAT-TM$\times$ N & - & 21.94         & 25.62         & -             & -    & 26.01       & 27.6$\times$\\
    TCL-NAT+NDP (s=9)    & NAT-TM$\times$ N & - & 25.37         & 29.60         & -             & -    & 29.30       & 16.0$\times$\\
    CMLM-small (iter=1)  & NAT-TM$\times$ N & - & 15.06         & 19.26         & 20.12         & 20.36         & -             & -\\
    CMLM-small (iter=10) & NAT-TM$\times$ N  & $\surd$ & 25.51         & 29.47         & 31.65         & \underline{32.27}         & -             & $<$1.5$\times$\\
    NAT-EM              & NAT-TM$\times$ N & - & 25.75          & 29.29        & -             & -    & -            & 9.14$\times$\\
    FCL-NAT  & NAT-TM$\times$ N & - & 21.70         & 25.32         & -             & -            & -            & 28.9$\times$\\
    FCL-NAT+NPD (s=9)  & NAT-TM$\times$ N & - & 25.75         & 29.50         & -             & -            & -            & 16.0$\times$\\
    NART-DCRF            & NAT-TM$\times$ N+DCRF$\times$ 1 & - & 23.44           & 27.22         & -             & -             & -             & 10.4$\times$\\
    NART-DCRF+LPD (s=9)   & NAT-TM$\times$ N+DCRF$\times$ 1 & - & 26.07       & 29.68         & -             & -             & -             & 6.14$\times$\\
    NART-DCRF+LPD (s=19)   & NAT-TM$\times$ N+DCRF$\times$ 1 & -  & \underline{26.80}       & 30.04         & -             & -             & -             & 4.39$\times$\\
    LevT                  & NAT-TM$\times$ N  & $\surd$    & \underline{27.27}       & -         & -      & \underline{33.26}      & -             & 4.01$\times$\\
    \midrule
    \bf{Our Models}\\
    \ReorderNAT (NAT)                             & NAT-TM$\times 1$  & -            &  22.79	      & 27.28          & 29.30         &   29.50       & 25.29       & 16.11$\times$\\
    \ReorderNAT (NAT)+LPD (s=7)                     & NAT-TM$\times 1$ & -             & 24.74    & 29.11	& 31.16     & 31.44   &  27.40   & $7.40\times$\\
    \ReorderNAT (AT)            & NAT-TM$\times N$-$1$ & -          & \textbf{26.49}	              & \textbf{31.13}              & \textbf{31.70}         & \textbf{31.99}               & 30.26 & 5.96$\times$\\
    \bottomrule
  \end{tabular}}
  \caption{Overall results of AT and NAT models in BLEU score on the test sets of WMT14 and WMT16, and validation set of IWSLT16. 
  ``AT-TM'' and ``NAT-TM'' denote the AT Transformer and NAT Transformer decoder block respectively. ``AT-GRU'' denotes the AT GRU decoder block. ``NAT-TM-imitate'' denotes the NAT Transformer decoder block with the imitation module. ``FlowStep'' denotes flow steps~\citep{ma2019flowseq}. ``DCRF'' denotes a CRF layer with dynamic transition~\citep{sun2019fast}. ``NPD'' denotes noisy parallel decoding~\citep{gu2018non}, ``LPD'' denotes length parallel decoding~\citet{wei2019imitation}, ``b'' denotes beam size and ``s'' denotes sample size. ``iter'' denotes translation refinement iterations. Better BLEU scores with {\em significant low speedup} are underlined.}
  \label{tab:overall-results:more}
\end{table*}

\begin{table*}
    \centering
    \small
    \begin{tabular}{lcccc}
    \toprule
    \multicolumn{1}{l}{\multirow{2}{*}{Model}} & \multicolumn{2}{c}{WMT14}     & \multicolumn{2}{c}{WMT16}\\
     & En$\to$De     & De$\to$En     & En$\to$Ro     & Ro$\to$En \\\midrule
    \ReorderNAT (NAT)	            & 21.96	& 27.19	& 29.55	& 30.62 \\
    \ReorderNAT (NAT) + LPD (s=7)	& 23.68	& 29.10	& 31.09	& 32.74 \\
    \ReorderNAT (AT)	            & 25.44	& 30.52	& 32.67	& 33.52\\
    \bottomrule
    \end{tabular}
    \caption{Results on the WMT14 En-De and WMT16 Ro-En validation sets.}
    \label{tab:overall-results:dev}
\end{table*}

\subsection{All Results on Chinese-English Translation}
We show all results on all datasets including MT02, MT03, MT04, MT05, MT06 and MT08 in Table~\ref{tab:chinese_english_all}.

\begin{table*}
  \centering
  \small
  \begin{tabular}{lcccccc}
    \toprule
    Model                   & MT02*      & MT03       & MT04   & MT05   & MT06   & MT08\\
    \midrule
    \multicolumn{3}{l}{\bf{Autoregressive Models}}\\
    Transformer$_{full}$    & 46.11	    & 43.74	    & 45.59	& 44.11 & 44.09	& 35.07\\
    Transformer$_{one}$     & 43.60	    & 41.24	    & 43.39	& 41.62 & 41.07 & 31.67\\
    Transformer$_{gru}$     & 43.68	    & 40.55	    & 43.02	& 40.73 & 40.32 & 31.09\\
    \midrule
    \multicolumn{3}{l}{\bf{Non-Autoregressive Models}}\\
    imitate-NAT             & 33.77     & 32.29     & 34.83   & 31.96 & 31.84	& 24.10\\
    imitate-NAT+LPD (s=7)         & 37.73	    & 36.53	    & 39.11   & 35.97 & 36.19 & 27.29\\
    \ReorderNAT (NAT)         & 37.99	    & 36.03	    & 38.17 & 36.07 & 36.28 & 27.99\\
    \ReorderNAT (NAT) + LPD (s=7)   & 41.58	    & 39.15	    & 41.67 & 39.71 & 39.58 & 30.44\\
    \ReorderNAT (AT)          & 45.22	    & 43.20	    & 44.89	& 43.45 & 42.96 & 	34.35\\
    \bottomrule
    \end{tabular}
    \caption{BLEU scores on Chinese-English translation. * indicates the validation set.}
    \label{tab:chinese_english_all}
\end{table*}

\subsection{Translation Quality v.s. Sentence Lengths}
Figure~\ref{fig:length} shows the BLEU scores of translations generated by the Transformer (AT model), the NAT model (\ReorderNAT without the reordering module), \ReorderNAT (NAT) and \ReorderNAT (AT) on the WMT14 and WMT16 test sets with respect to input sentence lengths.

\begin{figure*}[]
     \centering
     \subfigure[WMT14 En$\to$De]{
        \centering
        \includegraphics[width=0.48\columnwidth]{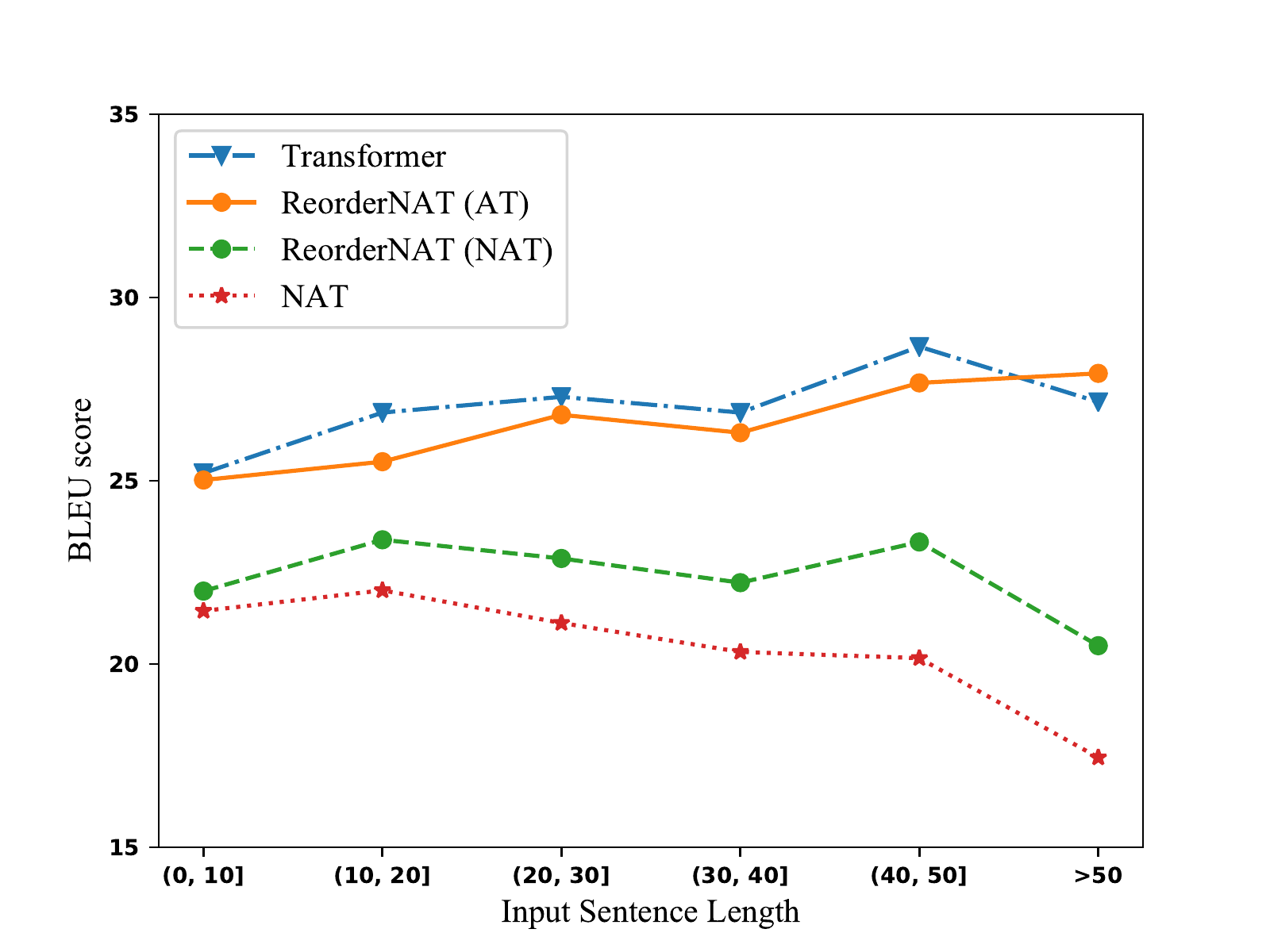}
     }
     \subfigure[WMT14 De$\to$En]{
        \centering
        \includegraphics[width=0.48\columnwidth]{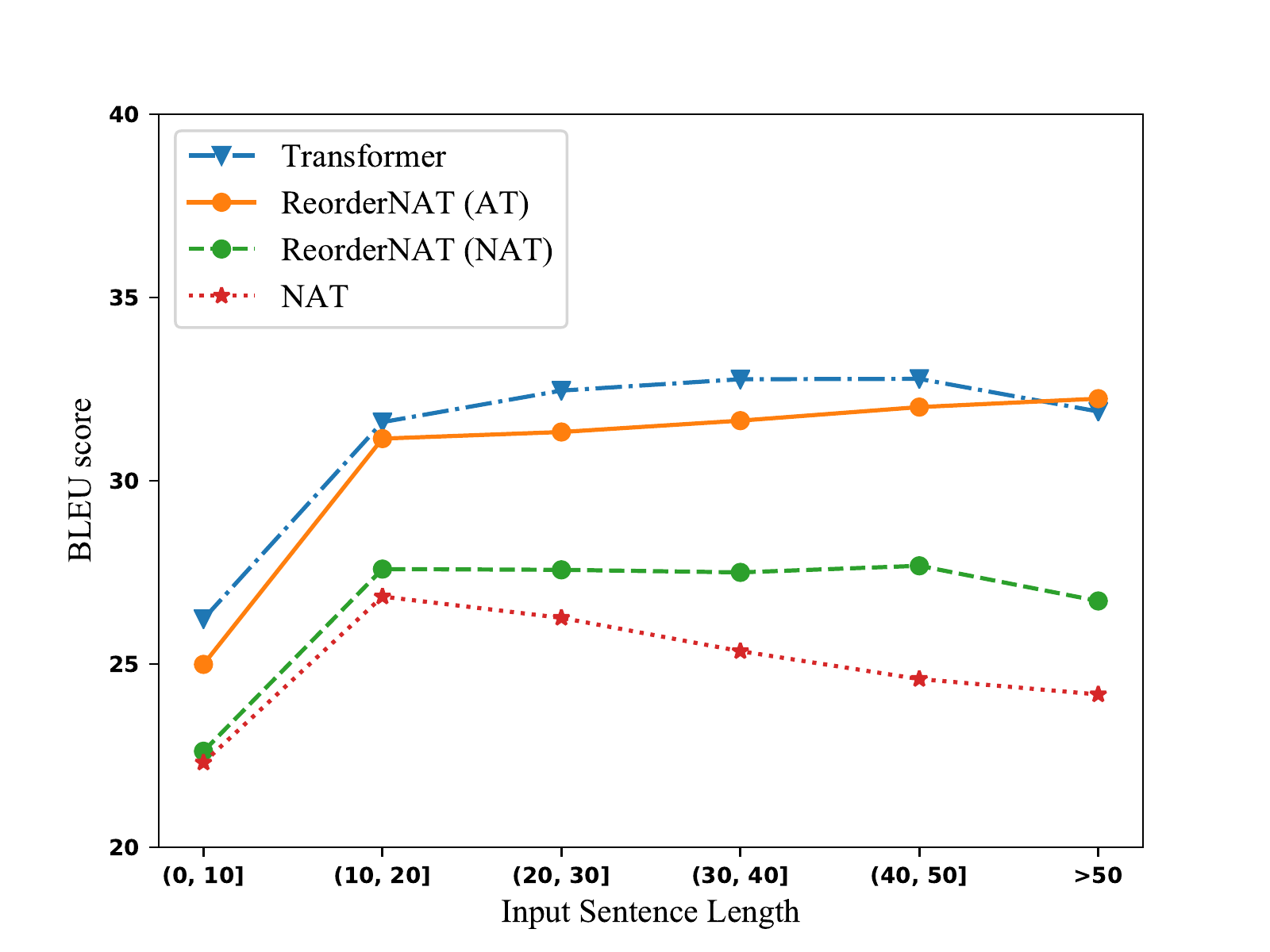}
     }
     \subfigure[WMT16 En$\to$Ro]{
        \centering
        \includegraphics[width=0.48\columnwidth]{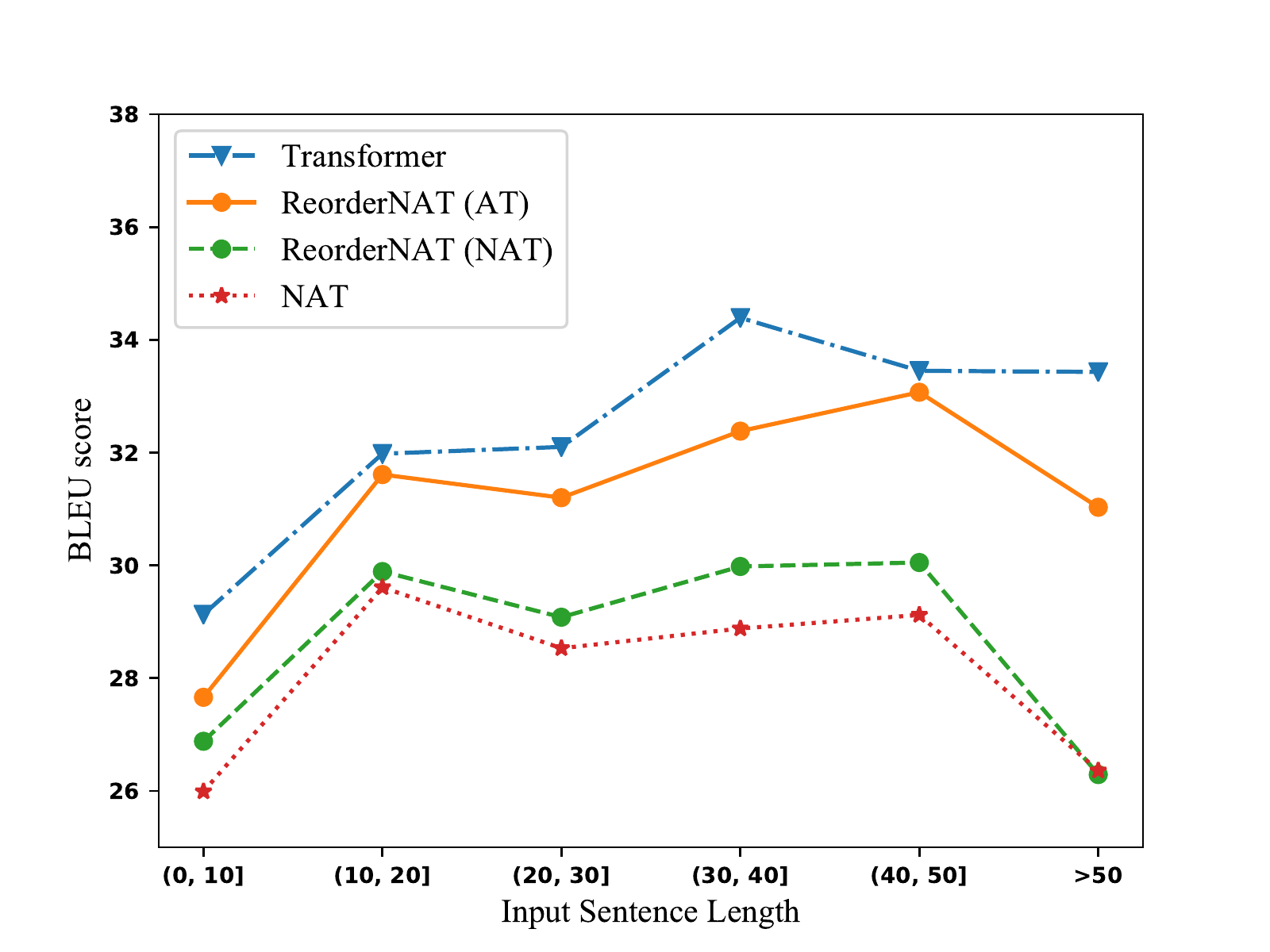}
     }
     \subfigure[WMT16 Ro$\to$En]{
        \centering
        \includegraphics[width=0.48\columnwidth]{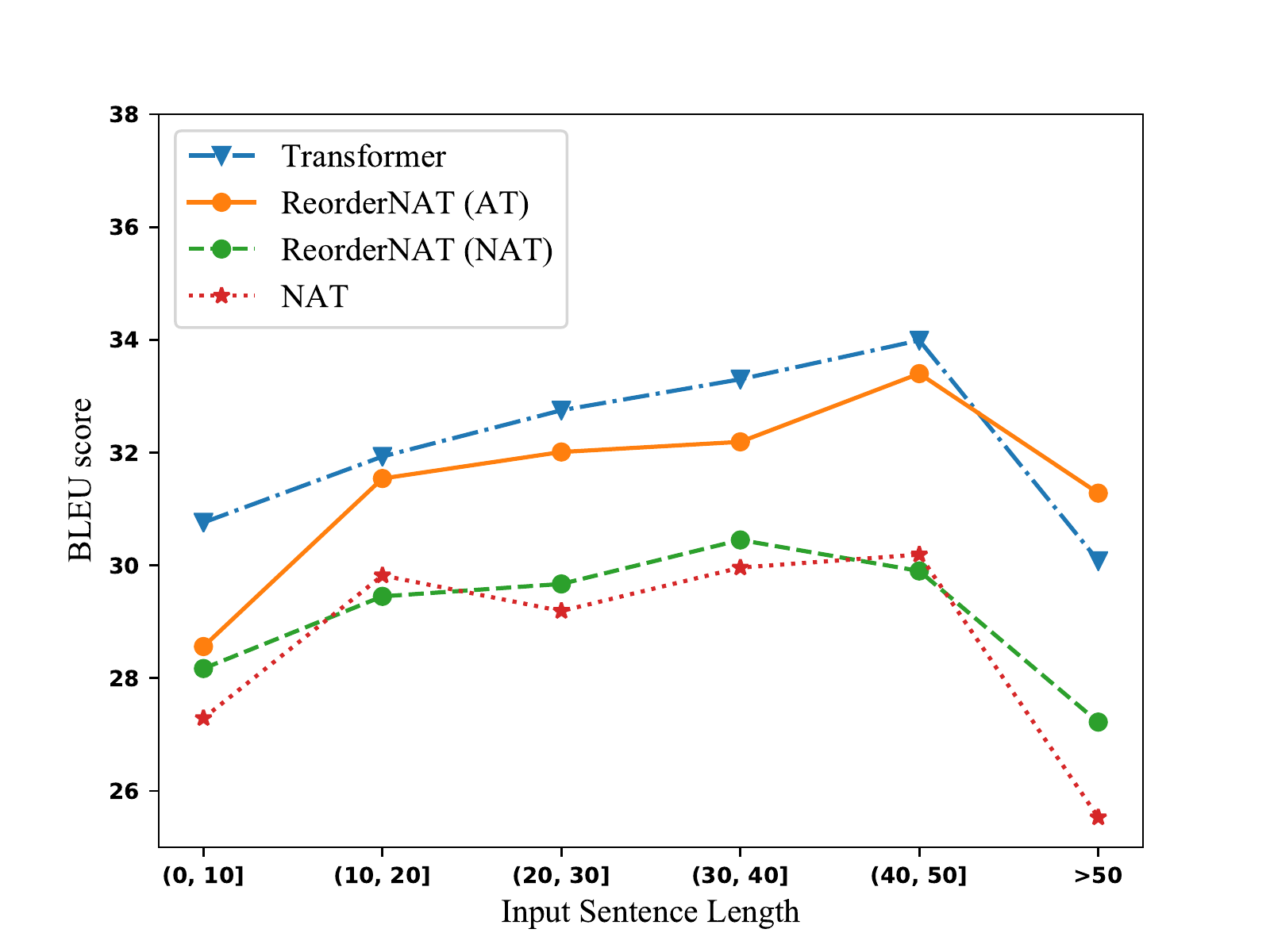}
     }
     \caption{Translation quality on the WMT14 and WMT 16 test sets over various input sentence lengths.}
     \label{fig:length} 
 \end{figure*}

\subsection{Multimodality Related Error Reduction}
We show an quantitative investigation of the reduction of the two most typical multimodality related errors, i.e., duplicate and missing words, on WMT14 En-De and WMT16 En-Ro tasks in Table~\ref{tab:multimodality-error}.

\begin{table*}
	\centering
	\small
	\begin{tabular}{lcrrrr}
		\toprule
		\multicolumn{1}{l}{\multirow{2}{*}{Model}} & \multicolumn{1}{l}{\multirow{2}{*}{Metric}}      &  \multicolumn{2}{c}{WMT14}     & \multicolumn{2}{c}{WMT16}  \\
		&   & En$\to$De     & De$\to$En     & En$\to$Ro     & Ro$\to$En \\
		\midrule
		NAT                           & Dup $\downarrow$    & 85.54     & 76.16     & 41.27     & 35.44\\
		\ReorderNAT (NAT)             & Dup $\downarrow$   & 18.83     & 15.67     & 28.12     & 24.85\\
		\ReorderNAT (NAT) + LPD (s=7) & Dup $\downarrow$   & 9.19     & 1.77     & 8.86      & 10.60\\
		\ReorderNAT (AT)              & Dup $\downarrow$   & 4.52      & -1.50     & 0.97      & 11.45\\
		\midrule
		NAT                           & Mis $\downarrow$   & 16.38     & 13.43     & 6.43      & 6.23\\
		\ReorderNAT (NAT)             & Mis $\downarrow$   & 12.76     & 9.53     & 4.97      & 5.07\\
		\ReorderNAT (NAT) + LPD (s=7) & Mis $\downarrow$   & 7.57      & 5.83      & 2.67      & 0.45\\
		\ReorderNAT (AT)              & Mis $\downarrow$   & 0.31      & 0.49      & 2.17      & 0.79\\
		\bottomrule
	\end{tabular}
	\caption{Relative increment of duplicate (``Dup'') and missing (``Mis'') token ratios on the WMT16 and WMT14 test sets. Smaller is better.}
	\label{tab:multimodality-error}
\end{table*}

\subsection{Effect of Guiding Decoding Strategy}
We show the results of \ReorderNAT (NAT) and \ReorderNAT (AT) using DGD and NDGD on all datasets in Table~\ref{tab:decoding_strategy_all}.

\begin{table*}
	\centering
	\small
	\begin{tabular}{lcccccccc}
		\toprule
		\multicolumn{1}{l}{\multirow{2}{*}{Model}} & \multicolumn{1}{l}{\multirow{2}{*}{Decoding Strategy}}      & IWSLT16   & \multicolumn{2}{c}{WMT14}     & \multicolumn{2}{c}{WMT16}  \\
		&               & En$\to$De & En$\to$De     & De$\to$En     & En$\to$Ro     & Ro$\to$En \\
		\midrule
		
		\multicolumn{1}{l}{\multirow{2}{*}{\ReorderNAT (NAT)}} & DGD  & 24.94 & 22.79         & 27.28         & 28.47         & 29.04\\
		& NDGD          & 25.29 &  21.05	      & 25.92          & 29.30         &   29.50\\
		\midrule
		\multicolumn{1}{l}{\multirow{2}{*}{\ReorderNAT (AT)}} & DGD   & 30.15  & {26.49}         & 31.17         & 31.52         & 31.95\\
		& NDGD          & {30.26} & {26.51}	              & {31.13}              & {31.70}         & {31.99}             \\
		\bottomrule
	\end{tabular}
	\caption{Effect of guiding decoding strategy.}
	\label{tab:decoding_strategy_all}
\end{table*}

\begin{table*}
  \centering
  \begin{tabular}{lcccccc}
    \toprule
     \% of training data &  &  10\%  & 30\%   & 50\%  & 70\%  & 100\%\\
    \midrule
    \multicolumn{1}{l}{\multirow{2}{*}{\ReorderNAT (NAT)}} & DGD & 24.48  &  24.58  & 24.92 & 25.07 & 24.94\\
    & NDGD & 25.19 & 24.85 & 25.22 & 25.47 & 25.29 \\
    \midrule
    \multicolumn{1}{l}{\multirow{2}{*}{\ReorderNAT (AT)}} & DGD  & 29.81  & 29.99   & 29.92 & 30.28 & 30.15\\
    & NDGD  & 29.93 &  30.28 & 29.86 & 30.01 & 30.26 \\
    \bottomrule
    \end{tabular}
    \caption{The BLEU scores on IWSLT16 validation set while varying the amount of training data for fast\_align.}
    \label{tab:alignment_quality}
\end{table*}

\subsection{Effect of Layer Number of Decoder Module}
\begin{figure}
	\centering
	\includegraphics[width=0.80\columnwidth]{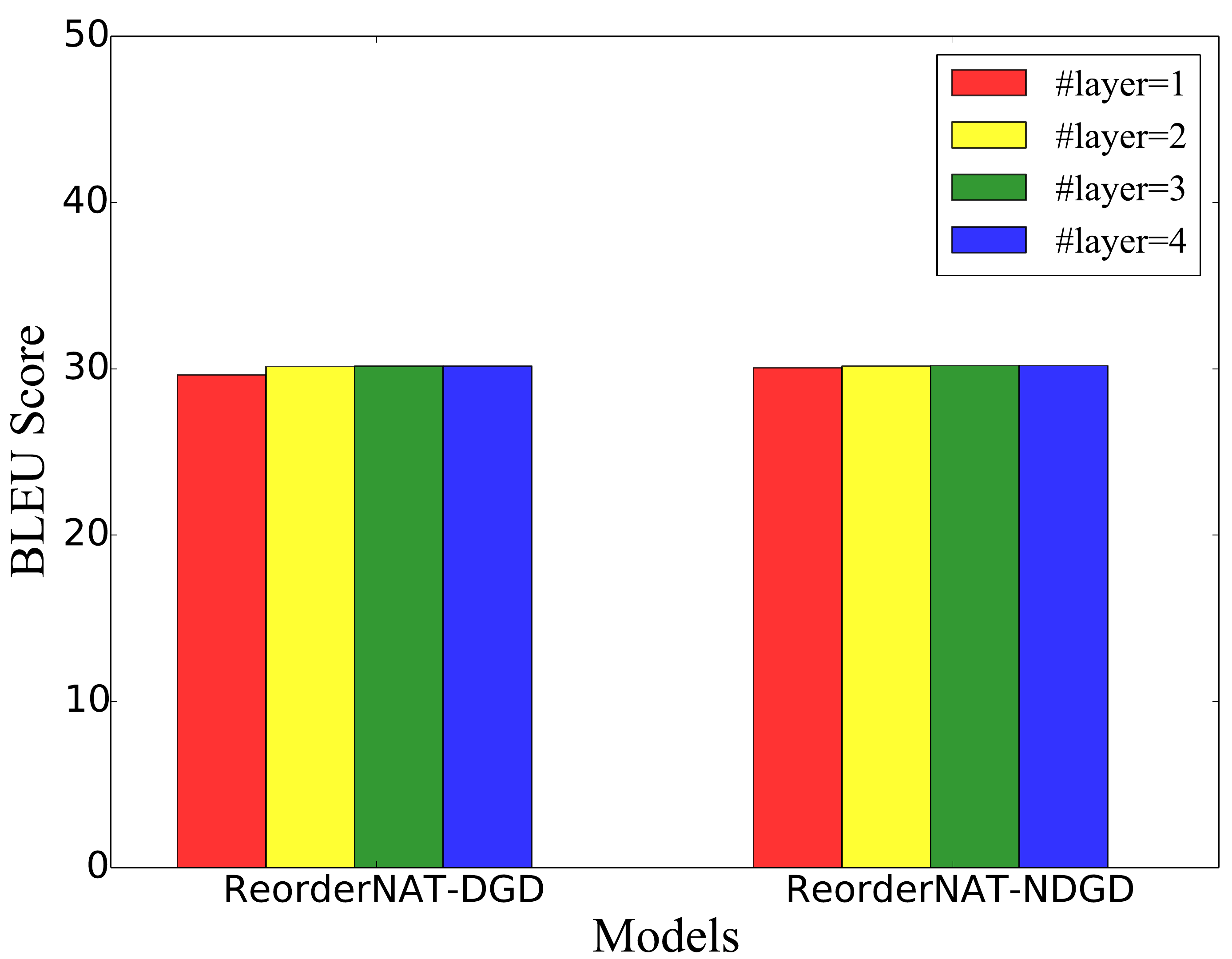}
	\caption{Effect of the Transformer decoder block number in the decoder module on the IWSLT16 validation set.}
	\label{fig:nat-layer} 
\end{figure}

To compare the NAT and AT baselines with the same model complexity, we utilize $N$-$1$ layers of Transformer decoder blocks for the decoder module in the experiments. However, we argue that the decoding space of decoder module is quite small and may be modeled by smaller architecture. To verify our assumption, we compare the translation quality of \ReorderNAT model with different Transformer layer numbers in decoder module on the IWSLT16 validation set. Figure \ref{fig:nat-layer} shows the results, from which we can see that our \ReorderNAT model also performs well with smaller decoder module, which demonstrates that our model can be further speedup.

\begin{table}[t!]
  \centering
  \small
  \begin{tabular}{lrrr}
    \toprule
    Model &  GPU & CPU \\
    \midrule
    Transformer$_{full}$                        & 1.00$\times$  &   1.00$\times$\\
    Transformer$_{one}$                         & 2.42$\times$  &   1.56$\times$\\
    Transformer$_{gru}$                         & 3.10$\times$  &   1.63$\times$\\
    \midrule
    \ReorderNAT (NAT)                           & 16.11$\times$ &  13.25$\times$\\
    \ReorderNAT (NAT)+LPD (s=7)                 & 7.40$\times$  &  1.46$\times$\\
    \ReorderNAT (AT)                            & 5.96$\times$  &  8.97$\times$\\
    \bottomrule
  \end{tabular}
  \caption{Speedup on CPU. As data for the other baselines are not available, we only show data for Transformers and our models.}
  \label{tab:cpu-speedup}
\end{table}
\begin{table}[t!]
    \centering
    \small
    \begin{tabular}{lcc}
    \toprule
    Model & WMT14/WMT16 & IWSLT16 \\\midrule
    \ReorderNAT (NAT)	& 45.78M     & 18.52M \\
    \ReorderNAT (AT)	& 47.04M	 & 18.75M \\
    \bottomrule
    \end{tabular}
    \caption{Parameter number.}
    \label{tab:model-size}
\end{table}

\subsection{Effect of Temperature Coefficient $T$}
\begin{figure}[!h]
	\centering
	\includegraphics[width=0.80\columnwidth]{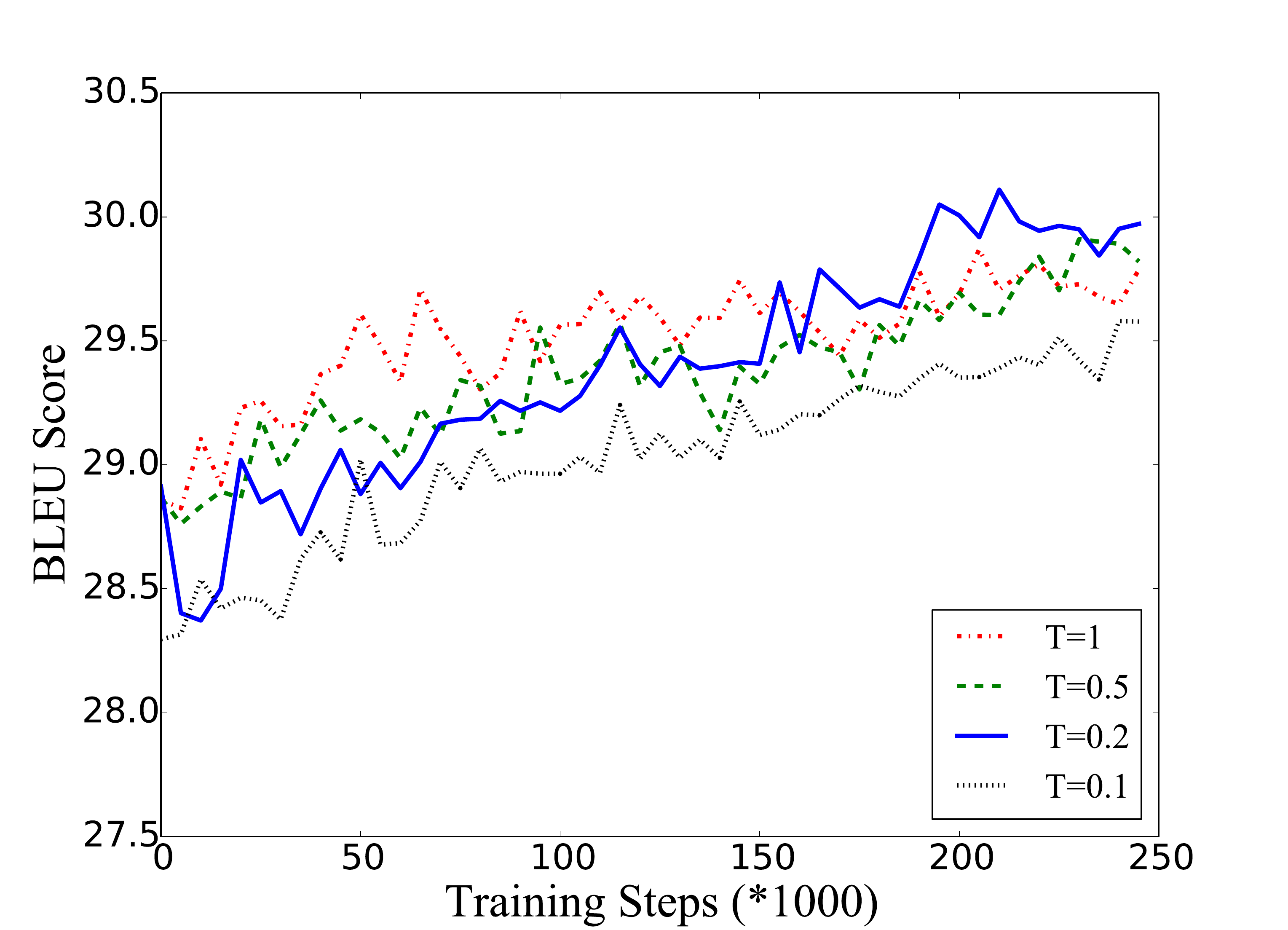}
	\caption{Effect of temperature coefficient $T$ on the IWSLT16 validation set.}
	\label{fig:T} 
\end{figure}

The temperature coefficient $T$ controls the smoothness of the $\mathbf{\mathcal{Q}}$ distribution (see Eq.~10 in the main text). As shown in Figure~\ref{fig:T}, we find that $T$ affects the BLEU scores on the IWSLT16 validation set to some extent. While $T=0.1$ deceases BLEU scores, $T=0.2$ improves translation quality significantly and consistently. However, increasing $T$ further to $0.5$ or $1$ results in worse translation quality compared to $T=0.2$ after training $150$k steps. Hence, we set $T=0.2$ for the NDGD strategy in our experiments. 

\subsection{Effect of Alignment Quality}

To investigate the effect of the alignment quality for training the pseudo-translation, we evaluate the performance of our \ReorderNAT model on IWSLT16 validation set when varying the amount of the training data used in fast\_align. The results are shown in Table~\ref{tab:alignment_quality}. From the table, we can observe that the BLEU scores are similar when using different proportions of training data for fast\_align, which indicates that our \ReorderNAT model is not sensitive to the alignment quality. Moreover, our \ReorderNAT model achieves higher BLEU scores on most proportions when using the NDGD strategy than using the DGD strategy, probably because the NDGD strategy could help mitigate the noise of pseudo-translation in decoding translations.

\subsection{Speedup on CPU}

The speedup of our \ReorderNAT model and AT baselines is shown in Table~\ref{tab:cpu-speedup}. Mathematical operations such matrix multiplication on CPUs are not as effective as on GPUs. Therefore, the speedup on CPU is smaller than that on GPU. One exception is that for our \ReorderNAT (AT) model. We believe the reason is that the AT reordering module is small such that it can be also computed effectively on CPU. Taking results in Table~\ref{tab:overall-results:more} into consideration, we conclude that \ReorderNAT (AT) should be a better choice for CPU as it achieves a better trade-off between speedup and translation quality.

\subsection{Parameter Number}

The parameter numbers of our models are shown in Table~\ref{tab:model-size}, which are comparable with those of the baselines.

\end{document}